\documentclass[twocolumn]{cinc}
\usepackage{amsmath, graphicx, hyperref}
\DeclareUnicodeCharacter{202F}{\,}

\title{ECG Classification with a Convolutional Recurrent Neural Network}
\author{Halla Sigurthorsdottir\textsuperscript{1}, Jérôme Van Zaen\textsuperscript{1}, Ricard Delgado-Gonzalo\textsuperscript{1}, Mathieu Lemay\textsuperscript{1} \\ \ \\
\textsuperscript{1}Swiss Center for Electronics and Microtechnology (CSEM), Neuchâtel, Switzerland }

\begin{document}
\maketitle

\begin{abstract}
We developed a convolutional recurrent neural network to classify 12-lead ECG signals for the challenge of PhysioNet/Computing in Cardiology 2020 as team Pink Irish Hat. The model combines convolutional and recurrent layers, takes sliding windows of ECG signals as input and yields the probability of each class as output. The convolutional part extracts features from each sliding window. The bi-directional gated recurrent unit (GRU) layer and an attention layer aggregate these features from all windows into a single feature vector. Finally, a dense layer outputs class probabilities. The final decision is made using test time augmentation (TTA) and an optimized decision threshold. Several hyperparameters of our architecture were optimized, the most important of which turned out to be the choice of optimizer and the number of filters per convolutional layer. Our network achieved a challenge score of 0.511 on the hidden validation set and 0.167 on the full hidden test set, ranking us 24rd out of 41 in the official ranking. 
\end{abstract}

\section{Introduction}\label{sec:introduction}
Cardiovascular diseases are both grave and prevalent and are the global leading cause of death~\cite{benjamin2019heart}. To help reduce the death rate of different types of cardiovascular diseases readily available, fast and accurate screening and early diagnosis is key. The goal of the PhysioNet/Computing in Cardiology Challenge 2020 is to develop open-source algorithmic approaches for detecting abnormalities from 12-lead ECG recordings~\cite{goldberger2000physiobank, PhysioNet2020}.

Our team, Pink Irish Hat, employed a similar neural network architecture to~\cite{chen2020detection}, which combines convolutional and recurrent layers, taking sliding windows of ECG signals as input and yielding the probability of each class as output. To get a final decision, we used test time augmentation (TTA) as well as an optimized threshold on the probabilities. We optimized several hyperparameters of the network and the training process as well as testing several methods to alleviate the class imbalance in the training set.

\section{Methods}\label{sec:methods}
This section includes label strategy, data pre-processing, network architecture, and hyperparameter optimization.
\subsection{Label Strategy}\label{sec:label}
Since only 27 out of the 111 diagnoses found in the dataset were scored in the challenge, we grouped all non-scored diagnoses into one ``negative'' class. Furthermore, three pairs of diagnoses were considered as equivalent in the challenge. We mapped all recordings labelled SVPB or VPB to PAC or PVC respectively, since their ECG patterns are similar enough not to confuse the network during training. However, since the QRS duration is different between CRBB and RBBB by definition, we did not combine these two diagnoses into one class.

\begin{figure*}[th!]
\centering
\includegraphics[trim={12cm 42cm 4cm 45cm},clip, width=\textwidth]{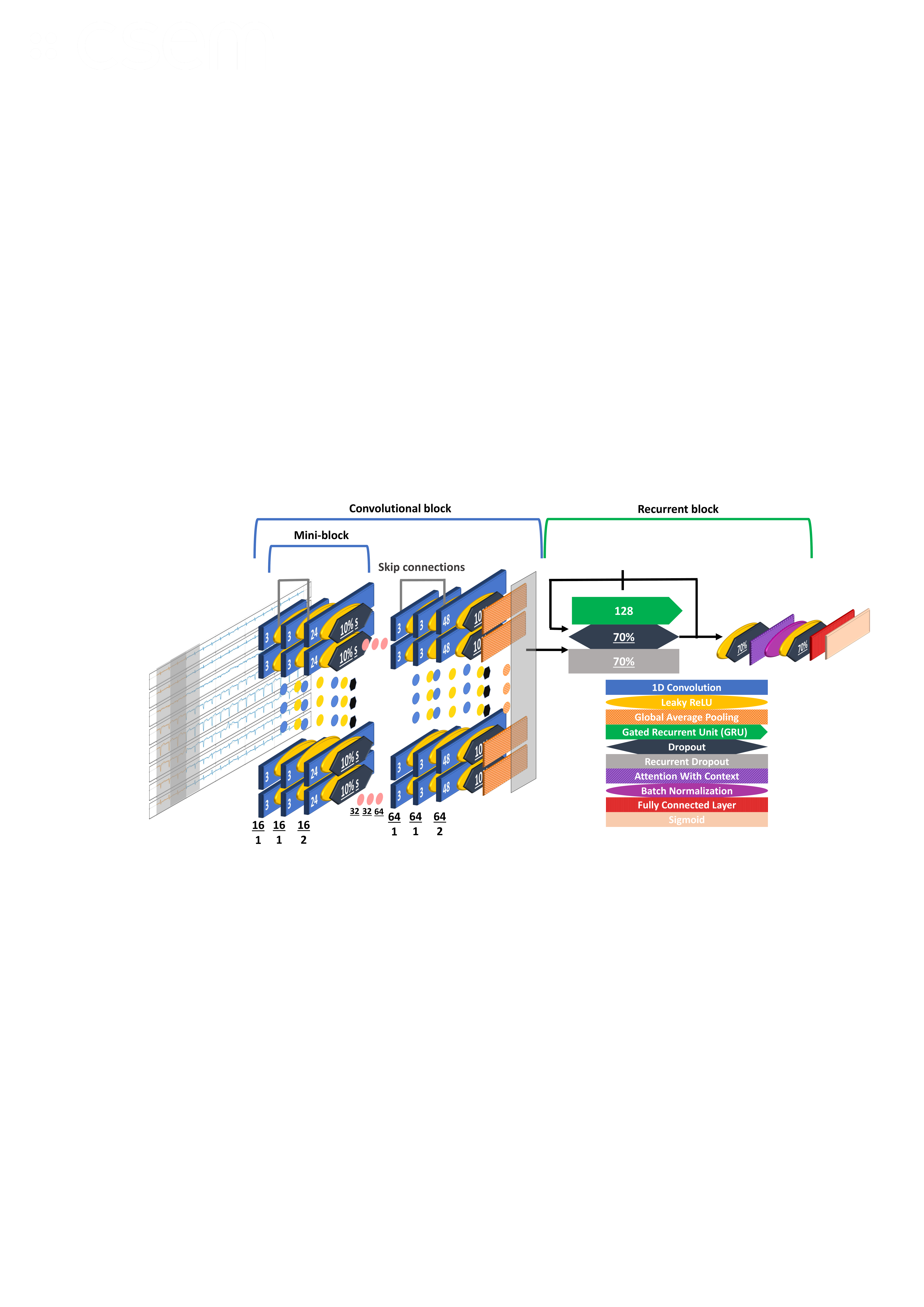}
\caption{The network architecture. The different layers are distinguished by color/shape/pattern as detailed in the legend. The network has a convolutional and a recurrent block. The \textbf{convolutional block} is made up of five smaller ``mini-blocks''. Each mini-block has three convolutional layers, each with a rectified linear unit (ReLU) activation. Printed on each convolutional layer is the filter size. Below each layer, there are the number of filters (upper) and stride (lower). Each mini-block is concluded with a dropout layer. In the first mini-block, the dropout layer is replaced by a SpatialDropout layer (denoted with an s). The dropout rate is printed on the layer. The gray arrows on each mini-block represent skip connections. Only the first and last mini-blocks are depicted in the figure. The other three are depicted as pink dots. The middle mini-blocks are the same as the last block with the exception that the kernel size of the last convolutional layer is the same as the first mini-block. Furthermore, the number of filters per convolutional layer are different. The number of filters of each mini-block (constant throughout the block) are printed below each pink dot. The \textbf{recurrent block} is made up of a gated recurrent unit (GRU) with both input and recurrent dropout, a ReLU activation and a dropout of the output, and an attention layer with context. The number of hidden units is printed on the GRU layer. The final layers are the batch normalization, ReLU activation, dropout, and a fully connected layer with a sigmoid activation which returns the non-exclusive class probabilities. Underlined values were considered as hyperparameters, their optimized values are found in the figure. More details on their optimization are found in Section~\ref{sec:hyperparameters} and Section~\ref{sec:hyperparameters_results}.}
\label{fig:architecture}
\end{figure*}

\subsection{Pre-processing and Augmentation}\label{sec:preproc}
We excluded leads with redundant information, namely leads III, aVR, aVL and aVF. Since the dataset includes recordings sampled at 257~Hz, 500~Hz and 1000~Hz, we resampled the data to the lowest of the three, 257~Hz. Since some of the very long recordings (30 minutes) have a significant drift, we applied a high pass filter with a cutoff at 0.5~Hz to all signals. We then split the data into training (80\%) and validation (20\%) [hereafter referred to as ``local validation''] sets stratified by dataset and the least common label found in each recording. Finally, each lead was normalized by the standard deviation over the training data. To augment the data and increase model generalization capabilities, a random offset was sampled for the beginning of the first window every time an ECG recording was presented to the network.

\subsection{Network Architecture and Training}\label{sec:architecture}
We selected a network architecture combining convolutional and recurrent blocks (see Figure~\ref{fig:architecture}). The convolutional block was used to extract features while the recurrent block was used to handle signals of varying length.
The convolutional block has 15 convolutional layers that are split into five mini-blocks. Each mini-block has three convolutional layers, each with a rectified linear unit (ReLU) activation function. The first two convolutions in the mini-block have a kernel size of 3 and stride 1, while the last has a kernel size of 24 and stride 2, excluding the very last convolution layer, which has a kernel size of 48. A dropout layer with a dropout rate of 10\% is added after each mini-block. In addition, each mini-block has skip connections. In all mini-blocks except the first, the skip connection is from the very beginning of the block, until before the last convolutional layer in the block. The first block is slightly different with the skip connections starting after the first convolution, but before the activation. This is to ensure that the input and output of the skip connection have the same dimensions. The convolutional block is concluded with a global average pooling layer which produces a feature vector for each window. These feature vectors are then fed into the recurrent block, composed of a bi-directional gated recurrent unit (GRU), with dropout for both the inputs and the recurrent state, followed by a ReLU activation, a dropout on the outputs and concluding with an attention layer with context~\cite{yang2016hierarchical}. The recurrent block produces one feature vector for the whole signal that is normalized over the batch and finally a fully connected layer with a sigmoid activation function produces the non-exclusive class probabilities. The final model was trained for 350 epochs, with a batch size of 8 recordings and the loss function was binary cross-entropy. Recordings with similar durations were grouped together to limit zero padding. The biggest batch size that did not exceed the hardware limitations during training was 8. The epoch with the highest local validation score was selected as the final model. The choice of optimizer was considered as a hyperparameter (see Section~\ref{sec:hyperparameters} and Section~\ref{sec:hyperparameters_results}). 
The model was implemented in TensorFlow~2.2.0.

\subsection{Test Time Augmentation}\label{sec:postproc}
To boost the performance of the model and to ensure that the offset of the input signal does not play a decisive role in the decision, we applied test time augmentation (TTA). To perform TTA, we applied ten different offsets to the input signal. A prediction was then made on each of these ECG signals. The mean of the resulting prediction probabilities was then thresholded to get the final decision. 

\subsection{Hyperparameter Optimization}\label{sec:hyperparameters}
Several hyperparameters of the network were optimized. To increase efficiency, the data was downsampled to 100~Hz and recordings longer than 200~s were excluded during hyperparameter optimization. Where possible, the model with the highest local  validation score was selected.

\textbf{Optimizer:} The first step was to find a suitable optimizer to minimize the binary cross-entropy loss. The tested optimizers were: Stochastic Gradient Descent (SGD),  Adam~\cite{kingma2017adam},  AMSGrad~\cite{reddi2019convergence}, and Nesterov Adam (Nadam)~\cite{dozat2016incorporating}. The optimizers were executed for 150 epochs with their default parameters in TensorFlow~2.2.0 on the full dataset resampled at 500~Hz optimizing the network from~\cite{chen2020detection} with an added global average pooling layer at the end of the convolutional block.

\textbf{Number of Filters:} The number of filters in~\cite{chen2020detection} was constant for all layers (12 filters per layer). We performed a grid search of this hyperparameter with the values of 12, 16, 32, 64, 128, and 256. Since models with 64 or more filters per layer exceeded the hardware limitations set by the challenge, a model with 16 filters in the first mini-block, 32 filters in the next two and 64 in the last two, which did not exceed the hardware limitations, was also tested.

\textbf{Dropout:} Two different dropout rates were used: one for the convolutional block ($d_c$) and one for the recurrent block ($d_r$). A grid search was run over these parameters. The range of the convolutional dropout was $d_c\in \left[0, 0.2\right]$ with a 0.1 increment. The tested dense/recurrent dropout rates were $d_c\in\{0.2, 0.5, 0.7\}$. 

\textbf{Decision Threshold:} Decision threshold optimization was performed after training by simply taking the prediction probabilities of the model on the local validation set and compute the challenge metric with different thresholds ranging from 0 to 1 with 0.1 increment.

Other hyperparameters tested either did not have an effect on the overall score (halving the kernel sizes to account for lower sampling rate, increasing number of mini-blocks by one), a negative effect (changing which layers within each mini-block would be followed by a dropout), or exceeded hardware limitations (\textit{e.g.}, additional GRU layer, doubling GRU number of hidden units). Moreover, additional methods to alleviate class imbalance were tested: focal loss~\cite{lin2017focal}, oversampling, MLSMOTE~\cite{charte2015mlsmote}, and an ensemble of binary and multi-label deep models. However, none of them successfully balanced the label distribution nor resulted in a higher local nor hidden validation score.

\section{Results}\label{sec:results}
In this section, we show the results of the hyperparameter optimization as well as the final hidden validation and testing scores.

\subsection{Hyperparameter Optimization}\label{sec:hyperparameters_results}
\begin{figure}[h!]
\centering
\includegraphics[width=\linewidth]{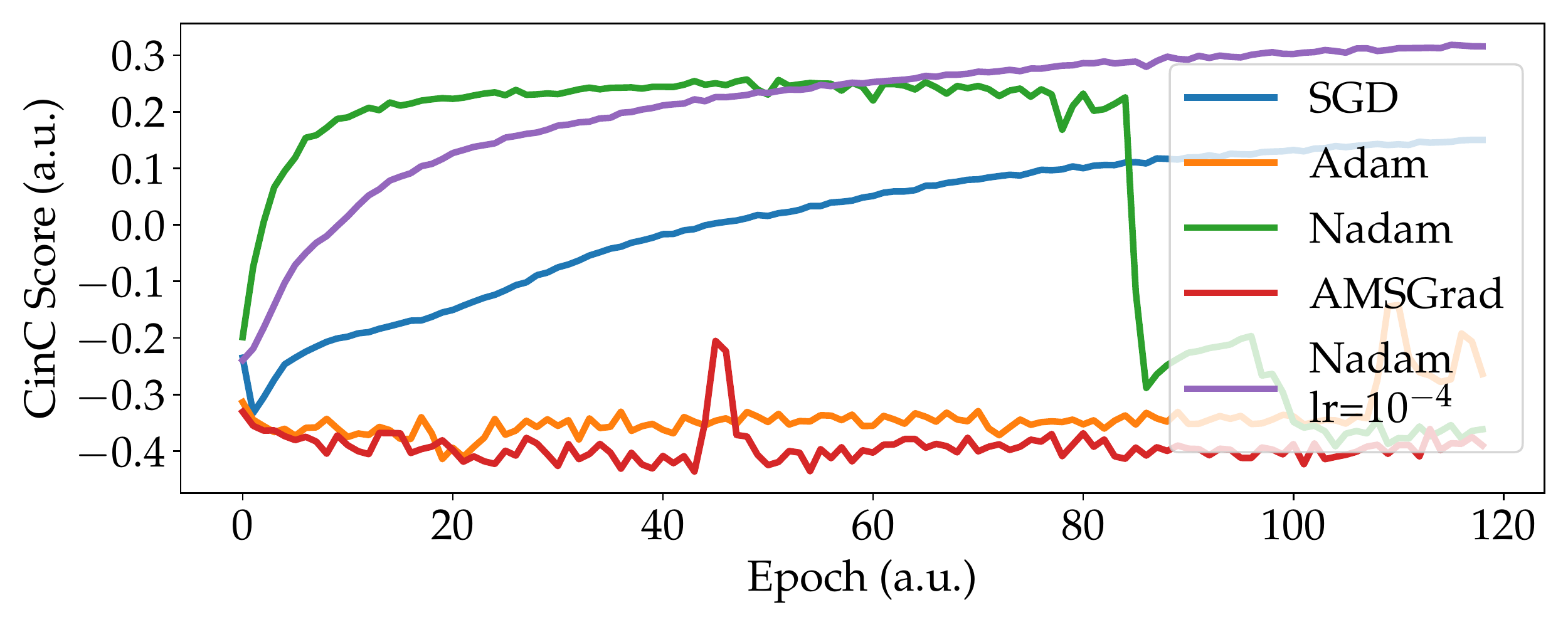}
\caption{The training scores over epochs obtained using different optimizers to optimize the model from~\protect\cite{chen2020detection} with an additional global average pooling layer.}
\label{fig:optimizers}
\end{figure}

\textbf{Optimizer:}
The training score over each epoch of each optimizer can be seen in Figure~\ref{fig:optimizers}. Nadam with the default parameters showed the most promise, but it was unstable. Therefore, we reduced its learning rate to $10^{-4}$. The local validation score showed a similar progression over epochs.

\begin{figure}[tpb]
\centering
\includegraphics[trim={4cm 3.5cm 3.5cm 2cm},clip, width=\linewidth]{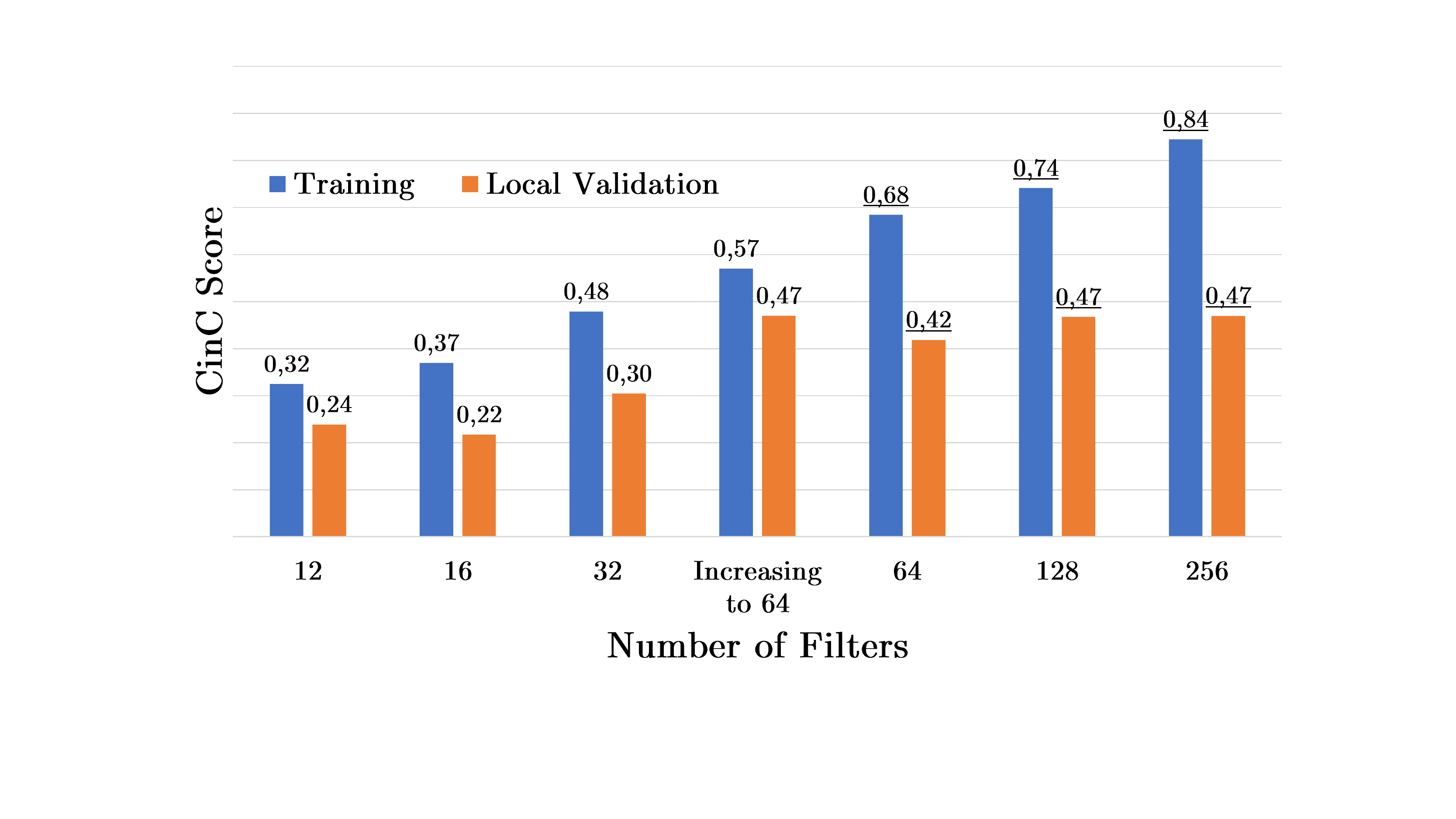}
\caption{The challenge score as a function of number of filters per layer. The model marked ``increasing to 64'' has 16 layers in the first mini-block, 32 in the next two, and 64 in the last two mini-blocks. Models with underlined scores exceeded the hardware limitations of the challenge.}
\label{fig:filters}
\end{figure}

\textbf{Number of Filters:} Training and local validation scores are depicted in Figure~\ref{fig:filters}. For a fair comparison, the scores are shown from the same epoch of all models. With more filters, the local validation performance goes up until it plateaus at 128 filters while the training performance goes up, a clear sign of overfitting. The models whose scores are underlined exceeded the hardware limitations. Note also that the local validation performance of the model with increasing number of filters with depth has a higher local validation score but lower training score than the equivalent model with constant number of filters, a sign of a more generalizable model.

\textbf{Dropout:} The local validation scores obtained were in the range $\left[0.38, 0.47\right]$. The combination of dropout rates with the highest local validation score of was $d_c=0.1$ and $d_r=0.7$. All scores mentioned are from the same epoch.

\textbf{Decision Threshold:} The best threshold for the model with the hyperparameters described in Section~\ref{sec:architecture} was 0.3 and yielded a local validation score of 0.573, a 0.027 improvement from the model with the default threshold (0.5).

Finally, applying TTA resulted in an increase in the range of $\left[0.0002, 0.02\right]$ depending on the model. The network was tested with and without skip connections, adding the skip connections resulted in an 0.1 increase in score.

\subsection{Final Results}\label{sec:hyperparameters_results}
The model with optimized hyperparameters yielded a score of 0.451 on the training set and 0.522 on the local validation set which increased to 0.546 with addition of TTA and up to 0.573 with the optimized threshold as well. The model with optimized hyperparameters and TTA was submitted with a threshold of 0.5 and with the optimized threshold of 0.3. The obtained hidden validation scores were 0.313 and 0.511 respectively. The model with a threshold of 0.3 was selected for testing on the final hidden test set resulting in a final test score of 0.167.

\section{Discussion and Conclusions}\label{sec:discussion}
Several models based on the convolutional-recurrent architecture were tested in an effort to obtain a model with good predictive capabilities with some rather good results. Moreover, several methods were investigated to cope with the class imbalance problem, but none of them improved model performance. Hardware limitations set by the challenge were an additional constraint for our deep neural network model working on long recordings, with the models that performed best on the local validation set exceeding these hardware limitations. Our best submitted model that respected the hardware limitations of the challenge achieved a score of 0.511 on the hidden validation set with threshold optimization and test time augmentation (TTA). Notice that the difference between the test score of the model with a decision threshold of 0.3 and 0.5 is approximately 0.2, which is very big. This was not expected, since the increase in local validation score when the threshold was optimized was an order of magnitude lower.

We observed that model generalization, class imbalance and the weak labelling of long recordings were the most important roadblocks to increase the performance of the overall system. Furthermore, we considered the challenge data to be uniformly labeled, but since the data comes from different sources, this assumption might not hold, leading to wrongly labeled signals. Advances in these areas will eventually lead to clinical decision support systems and monitoring devices with doctor-like accuracy.

\bibliographystyle{cinc}
\bibliography{references}

\begin{correspondence}
Halla Sigurthorsdottir\\
Rue Jaquet-Droz 1, Neuchâtel, NE, Switzerland\\
halla.sigurthorsdottir@csem.ch
\end{correspondence}

\balance

\end{document}